%% file: paper.tex
\documentclass{article}
\usepackage{lineno}
\usepackage[utf8]{inputenc}
\usepackage{color, soul}
\usepackage{natbib}
\usepackage{appendix}
\usepackage{booktabs}
\usepackage{etoolbox}
\usepackage{amssymb}
\usepackage[affil-it]{authblk}
\usepackage{graphicx}
\usepackage[labelfont=bf]{caption}
\usepackage{subfig}
\usepackage{algorithm}
\usepackage[noend]{algpseudocode}
\usepackage{fullpage}
\usepackage{authblk}
\usepackage{microtype}
\usepackage{graphicx}
\usepackage{booktabs}
\usepackage{hyperref}
\usepackage{amsmath}
\usepackage{mathtools}
\usepackage{amsthm}
\usepackage{color, soul}
\usepackage{subfig}
\usepackage[labelfont=bf]{caption}
\usepackage{listings}
\usepackage{xcolor}

\definecolor{codegreen}{rgb}{0,0.6,0}
\definecolor{codegray}{rgb}{0.5,0.5,0.5}
\definecolor{codepurple}{rgb}{0.58,0,0.82}
\definecolor{backcolour}{rgb}{0.95,0.95,0.92}

\lstdefinestyle{mystyle}{
    backgroundcolor=\color{backcolour},   
    commentstyle=\color{codegreen},
    keywordstyle=\color{magenta},
    numberstyle=\tiny\color{codegray},
    stringstyle=\color{codepurple},
    basicstyle=\ttfamily\footnotesize,
    breakatwhitespace=false,         
    breaklines=true,                 
    keepspaces=true,                 
    numbers=left,                    
    numbersep=5pt,                  
    showspaces=false,                
    showstringspaces=false,
    showtabs=false,                  
    tabsize=2
}

\setcitestyle{authoryear,round,citesep={;},aysep={,},yysep={;}}

\renewcommand{\cite}[1]{\citep{#1}}

\usepackage[capitalize,noabbrev]{cleveref}

\theoremstyle{plain}

\theoremstyle{definition}

\theoremstyle{remark}

\usepackage[textsize=tiny]{todonotes}

\newcommand{\lib}[1]{\texttt{#1}}
\newcommand{\mli}[1]{\mathit{#1}}
\DeclareMathOperator*{\argmax}{arg\,max}
\DeclareMathOperator*{\argmin}{arg\,min}

\newcommand{\piopt}{\pi_{\mli{Opt}}}
\newcommand{\picql}{\pi_{\mli{CQL}}}
\newcommand{\pioptfdpe}{\pi_{\mli{Opt-FDPE}}}
\newcommand{\pihist}{\pi_{\beta}}

\makeatletter
\renewcommand\AB@affilsepx{\protect\\ \protect\Affilfont}
\makeatother

\begin{document}

\newcommand{\splitatcommas}[1]{%
  \begingroup
  \begingroup\lccode`~=`, \lowercase{\endgroup
    \edef~{\mathchar\the\mathcode`, \penalty0 \noexpand\hspace{0pt plus 1em}}%
  }\mathcode`,="8000 #1%
  \endgroup
}

\title{Offline Deep Reinforcement Learning \\
            for Dynamic Pricing of Consumer Credit}

\author[1, 2, *]{Raad Khraishi}
\author[1]{Ramin Okhrati}

\affil[1]{\footnotesize Institute of Finance and Technology, UCL, London, United Kingdom}
\affil[2]{\footnotesize Data Science and Innovation, NatWest Group, London, United Kingdom}
\affil[*]{\footnotesize Corresponding author: Raad Khraishi, raad.khraishi.13@ucl.ac.uk}

\date{}

\providecommand{\keywords}[1]{\textbf{\textit{Keywords: }} #1}

\vskip 0.3in

\maketitle

\input{./sections/abstract.tex}

\vskip 0.2in
\input{./sections/introduction.tex}

\input{./sections/related-work.tex}
\input{./sections/methodology.tex}
\input{./sections/results.tex}
\input{./sections/conclusion.tex}
\input{./sections/acknowledgements.tex}

\bibliography{citations}
\bibliographystyle{plainnat}

\newpage
\appendix
\onecolumn

\input{./sections/appendix.tex}


\end{document}

%% file: sections/abstract.tex
\begin{abstract}
We introduce a method for pricing consumer credit using recent advances in offline deep reinforcement learning.
This approach relies on a static dataset and requires no assumptions on the functional form of demand.
Using both real and synthetic data on consumer credit applications,
we demonstrate that our approach using the conservative Q-Learning algorithm is capable of learning an effective personalized pricing policy without any online interaction or price experimentation.
\end{abstract}

%% file: sections/introduction.tex
\section{Introduction}
\label{section:introduction}

Consumer debt in the United States alone is worth over \$15 trillion.\footnote{\url{https://www.newyorkfed.org/medialibrary/interactives/householdcredit/data/pdf/HHDC_2021Q3.pdf}}
Despite the importance of this market, setting interest rates for debt products is done with varying levels of sophistication.
Two common techniques used by lenders today are risk-based and profit-based pricing \citep{phillips2020pricing}.
Risk-based pricing involves adding a fixed margin on top of the expected cost including default for a specific loan or pricing segment.
Profit-based pricing extends this by also incorporating the estimated responsiveness of customers or customer segments to price to find the profit-maximizing interest rate.

This present work builds on profit-based pricing by introducing a model-free reinforcement learning approach to finding optimal prices.
In particular, we develop an approach for pricing installment credit products such as mortgages as well as personal and student loans which form the bulk of the consumer debt market.
Setting prices for these products involves factoring in short-term rewards from the underwriting of the product as well as longer-term default risk, adverse selection, market competitiveness, and customer lifetime value. 
With reinforcement learning we are able to formulate this problem as a sequence of pricing decisions for each loan applicant.
This formulation helps induce long-term consequences of actions by allowing pricing decisions taken by the agent to affect the future state of the environment.
In addition, we use a model-free reinforcement learning algorithm in which an agent must learn to implicitly estimate individual price-responsiveness as it learns a pricing policy. This is a key departure from traditional profit-based pricing approaches that make strong assumptions on the functional form of demand.
Furthermore, a reinforcement learning approach also allows us to learn a dynamic pricing policy that can adapt to changes in behavioral patterns and the economy \cite{rana2014real}. 

Traditional reinforcement learning algorithms learning from scratch by pricing consumer loans in a live environment may start off by setting inaccurate prices that would incur significant financial and reputational costs for a lender.
The potential for these costs makes traditional online reinforcement learning difficult to apply in practice and motivates the need for offline reinforcement learning algorithms that are able to learn a policy from a static dataset with past pricing decisions and outcomes. 

Offline learning introduces several additional challenges.
In particular, the agent must assess counterfactual scenarios about what might happen following an action despite not having any examples of that behavior in the training set.
The decisions in the dataset used to learn a policy may be different than the decisions that it must learn to apply.
This problem of distributional shift leads traditional reinforcement learning approaches to overestimate the value of unseen outcomes and propagate poor decisions \citep{levine20}.
Given limited changes in prices historically, distributional shift and generalization are particularly problematic for pricing consumer credit products.
To mitigate these issues this paper uses the conservative Q-learning (CQL) algorithm to regularize Q-values and reduce overestimation of out-of-distribution actions \citep{kumar20}.

The main contribution of this paper is the introduction of an approach to pricing consumer credit that uses model-free offline deep reinforcement learning to learn a sequential pricing policy.
This approach makes no assumptions on demand often used in traditional pricing approaches that may introduce misspecification errors in unknown and non-stationary environments.
We also extend credit pricing to the full reinforcement learning problem in which current actions may impact the future state of the environment.
Using synthetic and real data on auto loans, we demonstrate that this approach is able to learn an effective policy using only a static dataset without any live interaction or price experimentation.

The rest of this paper is organized as follows.
\Cref{section:related-work} presents a review of related pricing and reinforcement learning literature.
\Cref{section:methodology} formulates credit pricing as a reinforcement learning problem and describes the algorithm and evaluation approaches used.
\Cref{section:results} presents our result on both historic and synthetic datasets of loan applications.
Finally, \cref{section:conclusion} presents our conclusions and possible extensions of our work.

%% file: sections/related-work.tex
\section{Related Work}
\label{section:related-work}

Despite many breakthroughs in reinforcement learning across several problem domains (see \citet{li2018deep} for an overview), the literature on pricing and  consumer credit remains sparse.
In perhaps the paper most closely related to our current work, \citet{trench2003} used tabular value-function iteration to choose credit line increases and interest rate decreases for credit cards at regular intervals.
They applied a model-based approach that required estimation of a separate static transition matrix and heavily discretized the state and action space to make the problem tractable.
In addition, they introduced several constraints on the action space such as only considering interest rate reductions or credit line increases.
Despite these simplifications, the authors state that their model was implemented by Bank One and estimated an associated increase in annual profit of approximately 5\% (equivalent to \$75 million per year).

We improve on their work in several ways.
First, we do not discretize prices or restrict actions to only price decreases.
Second, we use a model-free deep reinforcement learning approach with continuous prices and states that does not involve estimation of a separate transition matrix and does not assume knowledge of the demand curve.
In addition, we also treat credit pricing as a continuing infinite horizon problem which is better suited to capturing the delayed effects of adverse-selection and price competitiveness.

Several studies have applied profit-based pricing approaches to the same historical auto loans dataset used in \cref{section:results}. \citet{phillips2015effectiveness} implement a profit-based pricing approach using a logistic regression price-response model and a simplified profit objective on the same historical auto loans dataset used in this paper to estimate the effectiveness of field price discretion. 
\citet{ban2021personalized} extended this work by introducing an iterated approach to learning the price-response curve using a generalized linear regression model and the capacity for price experimentation.
While both papers cite substantial increases in revenue, 38\% and 47\% respectively, their estimated performance relies heavily on their assumed demand behavior.
In \cref{section:results}, we compare our reinforcement learning approach against a similar profit-based pricing approach.
Several bandit approaches have also been explored recently.
\citet{bastani2019meta} introduced an approach that uses Thompson sampling and a linear demand model to learn a pricing policy across multiple related products through pricing experiments.
They defined a discrete set of auto loan products whereas in our approach the number of products is not fixed and new products may be defined as part of the continuous state.
\citet{luo2021distribution} applied a contextual linear bandit approach using a modified linear upper confidence bound algorithm.  
As with \citet{ban2021personalized}, both these approaches rely on random and potentially costly price experimentation to learn a policy.
In addition, none of the papers extend credit pricing to the full reinforcement learning problem where actions may impact the future state of the environment and use simple linear function approximation.

Outside of consumer credit, many successful applications of dynamic pricing using bandit or reinforcement learning algorithms exist.
For example, \citet{cheung2017dynamic} developed pricing policies in the presence of unknown demand with limited ability to perform price experimentation and personalization.
They implemented their pricing strategy for an online deal website, Groupon, and estimated a 21\% increase in daily revenue.
\citet{chen2021statistical} explored personalized pricing as well as assortment optimization for airline seating reservations in a single-period problem.
\citet{cohen2020feature} developed an online contextual bandit approach for pricing online fashion products with each product defined by a set of features.
\citet{trovo2018} applied multi-armed bandit algorithms to online pricing of non-perishable goods in both stationary and non-stationary environments.
Several other papers have extended dynamic pricing to the full reinforcement learning problem where an agent must consider the long-term consequences of its actions.
For example, \citet{rana2014real} developed a model-free tabular Q-learning approach for selling a fixed inventory by a given deadline. 
Using simulated data, they found improved performance in the presence of demand misspecification using their model-free approach relative to classic parametrized pricing policies.
\citet{krasheninnikova2019} applied model-free reinforcement learning for finding a pricing policy for insurance renewals to maximize revenue subject to a minimum renewal rate constraint using a synthetic dataset.

Pricing consumer credit, however, has two unique characteristics that distinguish it from pricing many types of consumer goods and services.
First, adverse selection may arise when higher risk borrowers are less sensitive to price than borrowers with lower default risk \citep{phillips11}.
This behavior leads increases in prices to increase the riskiness of a lender's portfolio. Second, instalment loans are contractual products with terms that may vary from several months to several decades in which a borrower makes regular payments towards the original principal and interest.
As such, revenue is not realized directly upon purchase as future customer behavior, such as failing to repay or early repayment, over the term of the product may impact profitability.
Therefore, at the time the price of a loan is set, the loan's profitability is a random variable.

Offline reinforcement learning techniques have been applied to many different problems where learning online is costly or inefficient.
\citet{abe2010optimizing} implemented a constrained offline reinforcement learning approach based on Q-learning for optimizing debt collections using linear function approximation.
They also implemented their solution to manage collections for the New York State Department of Taxation and Finance and estimated expected savings of \$100 million over a three year period.
\citet{theocharous15} used a classic offline reinforcement learning algorithm, Fitted Q-Iteration \citep{ernst2005}, for personalized ad recommendations.
By using reinforcement learning methods, they found that their policy was able to take into account the long-term effect of ad choice on customer lifetime value.

%% file: sections/methodology.tex
\section{Methodology}
\label{section:methodology}

In this section we formulate credit pricing as a Markov decision process, introduce our offline reinforcement learning approach, and describe our approach to policy evaluation.

\subsection{Markov Decision Process}

Following \citet{levine20}, we define a Markov decision process to be a tuple $(\mathcal{S}, \mathcal{A}, T, d_{0}, r,  \gamma)$,
where $\mathcal{S}$ represents the set of states $s \in \mathcal{S}$,
$\mathcal{A}$ represents the set of actions $a \in \mathcal{A}$,
$T$ is the unknown transition dynamics $T(s_{t+1} | s_{t}, a_{t})$,
$d_{0}$ is the initial state distribution,
$r: \mathcal{S} \times \mathcal{A} \rightarrow \mathbb{R}$ denotes the reward function,
and $\gamma \in (0, 1)$ is the discount factor.
We also define credit pricing to be a continuing problem where each time step $t$ relates to a new application and the total number of time steps, $H$, equals $\infty$.

\paragraph{States.} Our state space, $\mathcal{S}$, includes information that is typically available in a loan application such as credit score, the type of loan, the term of the loan, the amount of the loan, and competitor rates.

For example, with the auto loans dataset described in \cref{section:data}, we define
$s_{t} = \splitatcommas{[\mli{Term}_{t}, \mli{Amount}_{t}, \mli{FICO}_{t}, \mli{PD}_{t}, \mli{PreviousRate}_{t}, \mli{CompetitionRate}_{t}, \mli{PrimeRate}_{t}, \mli{Tier}_{t}, \mli{LoanType}_{t}, \mli{CarType}_{t}, \mli{PartnerBin}_{t}, \mli{State}_{t}, \mli{Months}_{t}, \mli{DayOfWeek}_{t}, \mli{MonthOfYear}_{t}, \mli{DaysSinceApp}_{t}]}$. See \cref{appendix:data-dictionary} for more details.


Additional features that are typically available through a credit application process such as tenure, channel, debt obligations, and income can be used to extend the state space though care must be taken to ensure any feature does not lead to biased or unfair pricing policies.

Although we refer to state for simplicity, our environment is partially observable due to incomplete information on factors that may impact a lender's profitability.
For example, a potential borrower who has recently lost their job may be more willing to accept an uncompetitive interest rate for fear of not receiving another offer later.
This behavior would impact both the price sensitivity of the borrower as well as their riskiness which may induce adverse selection and is not captured by the feature set (for a detailed discussion of price-dependent risk see \citet{phillips11}).

\paragraph{Actions.}

Upon each application the agent chooses an interest rate, $a_{t}$, quoted as an annual percentage rate (APR) which determines the price of the loan.
We define our action space, $\mathcal{A}$, to be the subset of positive real numbers, i.e. $a_{t} \in \mathbb{R}^{+}$.
This may be extended to include product fees as well.

\paragraph{Rewards.}

Many alternative financial measures of profit such as Net Income or Net Interest Income may be used to define rewards.
For example, with the auto loans dataset, we use a simplified expected profit measure from \citet{phillips2015effectiveness} which accounts for interest income, capital costs, and credit risk.
We use the shorthand $r_{t} = r(s_{t}, a_{t})$ to denote a scalar reward value, which is defined by:

\begin{small}
\begin{equation}
    r(s_{t}, a_{t}) = p(\mli{Accept}_{t} | a_{t}, s_{t}){} *  
    \Bigg  [ \Bigg . (1 - \mli{PD}_{t}) * (\mli{TotalPayment}_{t} - \mli{CapitalCost}_{t}) 
   - \mli{PD}_{t} * \mli{LGD}_{t} * \mli{CapitalCost}_{t} \Bigg . \Bigg ]  \,,
\label{eqn:reward}
\end{equation}
\end{small}

where $p(\mli{Accept}_{t} | a_{t}, s_{t})$ denotes the probability of a potential borrower accepting a loan given a price, $a_{t}$, and state feature vector, $s_{t}$.
$PD_{t}$ is the estimated probability of default of the applicant.\footnote{
The probability of default, $PD$, is not directly available in the dataset and is estimated from the FICO scores using data from \citet{munkhdalai2019empirical} which covers roughly the same time period. However, as their data is not specific to auto loans we expect our $PD$ values to be conservative estimates of default risk.}
We set the loss given default, $LGD$, to 50\% for all customers as in \citet{phillips2015effectiveness}. $\mli{TotalPayment}_{t}$ is shorthand for the total value of interest and principle payments of the loan over the course of its term as a function of its interest rate, $a_{t}$, loan amount, and term.
A similar function is used for the cost of capital, $\mli{CapitalCost}_{t}$,  with the prime rate, $PrimeRate_{t}$, used in place of $a_{t}$.

The goal of our pricing algorithm will be to learn a policy $\pi(a_{t}|s_{t})$ that defines a distribution over actions conditioned on states.
We can use this to define a trajectory, $\tau$, to be a (potentially infinite) sequence of states and actions $(s_{0}, a_{0}, ..., s_{H}, a_{H})$, admitting the following distribution:

\begin{small}
\begin{align*}
    p_{\pi}(\tau) &= d_{0}(s_{0}) \prod_{t=0}^{H} \pi(a_{t} | s_{t}) T(s_{t+1}|s_{t}, a_{t}) \,.
\end{align*}
\end{small}

We may also define the return, $R_{t}$, to be sum of discounted rewards $\sum_{t=0}^{T} \gamma^{t}r(s_{t}, a_{t})$.
Our objective then becomes to find a policy that maximizes the expected return, $\mathbb{E}_{\tau \sim p_{\pi}(\tau)} \left [ \sum_{t=0}^{H} \gamma^{t}r(s_{t}, a_{t}) \right ]$, denoted by $J(\tau)$.

\subsection{Reinforcement Learning}

Traditional pricing techniques for consumer credit such as risk-based pricing fail to capture idiosyncrasies in customer behavior and responsiveness to prices \cite{phillips2020pricing}.
While more sophisticated profit-based pricing approaches attempt to estimate this they require strong assumptions on the functional form of demand \cite{phillips2015effectiveness, ban2021personalized}.
In addition, these techniques are often myopic - they use the estimated price-response curve to find the profit-maximizing prices without taking into account the long-term consequences of actions.
Reinforcement learning algorithms help address these issues by allowing us to train an agent to learn to make sequential decisions based on past experience in an unknown and non-stationary environment \cite{sutton2018reinforcement}.
In addition, model-free reinforcement learning techniques do not assume knowledge of the reward function, $r(s_{t}, a_{t})$, price-response function, $p(Accept_{t} | s_{t}, a_{t})$, or transition probabilities, $T(s_{t+1}|s_{t}, a_{t})$.

We use the value function, $V_{\pi}$, to refer to the estimated value of being in a state, $s_{t}$, and following policy $\pi(a_{t} | s_{t})$. We also define the action-value function, $Q_{\pi}$, to be the estimated value of being in state, $s_{t}$, taking action $a_{t}$ and then following policy $\pi$. These functions are defined as follows:

\begin{small}
\begin{align*}
    V_{\pi}(s_{t}) &= \mathbb{E}_{\tau \sim p_{\pi}(\tau | s_{t})} \left [ \sum_{t'=t}^{H} \gamma^{t' - t}r(s_{t}, a_{t}) \right ] \,, \\ 
    Q_{\pi}(s_{t}, a_{t}) &= \mathbb{E}_{\tau \sim p_{\pi}(\tau | s_{t}, a_{t})} \left [ \sum_{t'=t}^{H} \gamma^{t' - t}r(s_{t}, a_{t}) \right ]  \,.
\end{align*}
\end{small}

Writing the  action-value function only in terms of $Q$:

\begin{small}
\begin{equation*}
    Q_{\pi}(s_{t}, a_{t}) = r(s_{t}, a_{t}){} +
         \gamma \mathbb{E}_{s_{t + 1} \sim T(s_{t + 1} | s_{t}, a_{t}), a_{t+1} \sim \pi(a_{t+1}|s_{t+1})} [Q_{\pi}(s_{t+1}, a_{t+1})]\,,
\end{equation*}
\end{small}

gives rise to the Q-learning algorithm which iteratively applies the Bellman optimality operator, $\mathcal{B}^{*} Q(s_{t}, a_{t}) = r(s_{t}, a_{t}) + \gamma \mathbb{E}_{s_{t+1} \sim T(s_{t+1} | s_{t}, a_{t}})[\max_{a_{t+1}
}Q(s_{t+1}, a_{t+1})]$, to find the optimal Q-function.
These estimates are then used to take actions usually following an implicit policy (e.g., $\epsilon{\text -}\mli{greedy}$).
The Q-values may be approximated using a deep neural network with parameters $\theta$, in which case we use $Q_{\theta}$ to denote the action-value function.

Actor-critic algorithms are another class of popular reinforcement learning algorithms and closely resemble classic policy iteration from dynamic programming \cite{sutton2018reinforcement}.
These algorithms directly estimate both a parameterized value function, $Q_{\theta}$, as well as a policy parameterized by $\phi$, $\pi_{\phi}$, by alternating policy evaluation and policy improvement steps:

\begin{small}
\begin{equation*}
\begin{aligned}
    \hat{Q}_{\theta}^{k+1} &\leftarrow \mathrm{arg} \min_{\theta} \mathbb{E}_{s_{t}, a_{t}, s_{t + 1} \sim \mathcal D}
    \Bigg [  \Bigg . ( (r(s_{t}, a_{t}) + \gamma \mathbb{E}_{a_{t+1} \sim \hat{\pi}_{\phi}^k(a_{t+1} | s_{t+1})}[\hat{Q}_{\theta}^{k}(s_{t+1}, a_{t+1})])  
   - Q(s_{t}, a_{t})  )^{2}  \Bigg . \Bigg ] \,, \\
    \hat{\pi}_{\phi}^{k+1} &\leftarrow \mathrm{arg} \max_{\phi} \mathbb{E}_{s_{t} \sim \mathcal{D}, a_{t} \sim \pi_{\phi}^{k}(a_{t} | s_{t})} \left [\hat{Q}_{\theta}^{k + 1}(s_{t}, a_{t}) \right ] \,,
\end{aligned}
\end{equation*}
\end{small}

where $\mathcal{D} = \{(s_{t}, a_{t}, r_{t}, s_{t + 1})\}$ refers to a dataset with states, actions, and rewards. The $\argmin$ and $\argmax$ in the equations are often approximated using gradient descent steps.

Standard Q-learning and actor-critic methods are off-policy algorithms that typically rely on some online interaction with the environment when learning.

\subsubsection{Offline Reinforcement Learning}

Learning to price consumer credit through trial-and-error using standard online reinforcement learning techniques can lead to costly mistakes at the detriment of both the lender and consumers.
More generally, organizations may also restrict the ability to perform any sort of live pricing experiments.
These restrictions on learning from live interaction motivate the need for training a pricing policy using a dataset of historical pricing decisions and outcomes.

In offline reinforcement learning, the dataset, $\mathcal{D}$, is gathered by a different policy than what our agent learns - in our case, a static dataset of historical pricing decisions.
To denote the behavioural policy used to collect the dataset we use  $\pi_{\beta}(a_{t} | s_{t})$. The goal of our reinforcement learning agent then becomes to learn a policy, $\pi$, from a dataset, $\mathcal{D}$, collected from behavioral policy $\pihist$, to maximize expected return, $J$, when interacting with the environment at test time.

While many online reinforcement learning techniques can be used to learn from an offline dataset they often suffer from distributional shift which results in poor performance \citep{levine20}.
Learning a policy from a static dataset requires assessing counterfactual actions that may not have been observed historically.
With the absence of corrective feedback, standard online techniques may end up over-estimating the value of unseen actions and states which in turn propagate further errors as the agent uses these inaccurate estimates at test time.

Offline (or batch) reinforcement learning algorithms mitigate these issues by attempting to learn to improve on the historical policy seen in the data while reducing deviation from historical actions.
This is often done through policy constraints that limit deviation from the historical policy, accounting for uncertainty in action-value estimates, or using conservative value function estimates that penalize out-of-distribution actions \citep{levine20}.
After initial training, many offline algorithms may be combined with online fine-tuning to boost performance.

\subsubsection{Conservative Q-Learning}

We use the conservative Q-learning algorithm (CQL) introduced by \citet{kumar20} for our pricing agent. CQL is a model-free offline reinforcement learning algorithm which has achieved state-of-the-art performance on a number of offline tasks.

CQL mitigates distributional shift by penalizing values for out-of-distribution state-action tuples during training.
In the implementation of CQL that we use this is done by adding an additional regularization term, $\mathbb{E}_{s \sim \mathcal{D}} [\log \sum_{a} \exp Q(s, a)]$, when learning the Q-function.
This helps mitigate distributional shift by pushing down Q-values for action-values sampled from the current policy as they are more likely to be out-of-distribution.
To avoid underestimation or producing estimates that are too conservative a second term is added to maximize the values of state-action tuples seen in the historical data, $ \mathbb{E}_{s, a \sim \mathcal{D}}[Q_{\theta_{i}}(s, a)]$. Combined these two terms produce a conservative Q-function that is a lower bound in expectation on the true value (\citet[Theorem 3.2]{kumar20}).

We use the soft actor-critic (SAC) \cite{haarnoja2018soft} version of the CQL algorithm implemented in the \lib{d3rlpy} package \cite{seno2020d3rlpy} and use $\picql$ to refer to the learned policy.
This version modifies the soft actor-critic loss to include the additional CQL regularization term:

\begin{small}
\begin{equation}
\label{eqn:cql_loss}
L(\theta_{i}) = \alpha \mathbb{E}_{s_{t} \sim \mathcal{D}} \Bigg [ \Bigg .  \log \sum_{a} \exp Q_{\theta_i}(s_{t}, a) 
              - \mathbb{E}_{s, a \sim \mathcal{D}}[Q_{\theta_{i}}(s, a)] - \kappa \Bigg . \Bigg ] + L_{SAC}(\theta_{i}) \,,
\end{equation}
\end{small}

where $L_{SAC}$ refers to the soft actor-critic loss without CQL regularization.
As in the original soft actor-critic paper, we use two critics parameterized by $\theta_{i}$ to speed up training and reduce positive bias in the policy improvement step \citep{haarnoja2018soft}.
The parameter $\alpha$ controls the degree of conservativeness and is automatically tuned by Lagrangian dual gradient descent.
$\kappa$ is a user-defined threshold value that helps control the magnitude of $\alpha$.
The $\log \sum_{a} \exp Q(s, a)$ term used to penalize Q-values is estimated using samples from the current policy:

\begin{small}
\begin{equation*}
    \log \sum_{a} \exp Q(s, a)  \approx \log \Bigg ( \Bigg . \frac{1}{2N} \sum_{a_{i} \sim \text{Unif}(a)}^{N} \left [ \frac{\exp Q(s, a_{i})}{\text{Unif}(a)} \right ]
    + \frac{1}{2N} \sum_{a_{i} \sim \pi_{\phi}(a|s)}^{N} \left [ \frac{\exp Q(s, a_{i})}{\pi_{\phi}(a_{i} | s)} \right ] \Bigg . \Bigg ) \,,
\end{equation*}
\end{small}

where $N$ refers to the user-defined number of sampled actions, and $\text{Unif}$ is the uniform distribution.

\subsection{Performance Evaluation}
\label{section:eval}

We are unable to test our policy through live interaction and instead evaluate our new approach using historical data.
To do so, we apply our model to an out-of-sample test set and predict performance using an estimated price-response model.
We compare our results against historical pricing decisions and a popular profit-based pricing approach.
In \cref{section:synthetic-results}, we also evaluate the performance of our policy on a synthetic dataset where the true demand function is known.

\subsubsection{Offline Policy Evaluation}
\label{section:ope}

Without knowledge of the true price-response function, $p(Accept_{t} | a_{t}, s_{t})$, or the transition probabilities, $T(s_{t+1}|s_{t}, a_{t})$, and without the ability to test our policy through live interaction it is difficult to assess model performance on the static auto loans data set.
Offline policy evaluation is an area of active research with a variety of different methods that include importance sampling, models of the transitions and rewards, and value-based estimators \cite{paine2020hyperparameter, voloshin2019empirical}.
Following \citet{phillips2015effectiveness, ban2021personalized, luo2021distribution}, we use a model-based approach and focus on estimating the price-response function  in \cref{eqn:reward} to predict the probability of a customer accepting a new price and use that to calculate expected reward.
Although a potentially significant advantage of our approach is in treating credit pricing as a sequential decision problem, we do not develop a model for the transition dynamics, $T(s_{t+1} | s_{t}, a_{t})$, to evaluate this.

Our primary price-response model is a logistic regression model trained on the full dataset, however, the choice of model and feature set used to estimate the price-response function impacts estimates of performance.
To account for the impact of misspecification we also estimate several alternative models using parametric and non-parametric techniques to provide a range of performance.

\subsubsection{Comparison}

Our first point of comparison is to the historical pricing policy, $\pi_{\beta}$, captured in the data.
Unfortunately, we could find no information on how pricing decisions were made by the lender at the time.
However, using the data we are able to directly calculate rewards and compare total expected reward from each of the policies over the test set, see the \cref{appendix:train-split}.
In addition, we also measure the mean absolute percentage deviation (MAPD) in prices from the historical policy to capture how different the pricing policies are.

Our second comparison is to profit-based price optimization, an approach commonly used in practice \cite{phillips2020pricing, ban2021personalized, besbes2009dynamic, phillips2015effectiveness}.
At a high-level, this approach involves two steps.
We first estimate a price-response model to estimate how likely a potential borrower is to take out a loan at a given price point,  $p(Accept_{t} | a_{t}, s_{t})$.
This price-response model is then used to find the greedy price that maximizes the reward function, $\max_{a_{t}} r(s_{t}, a_{t})$, defined in \cref{eqn:reward}.
We use $\piopt$ to denote this profit-based price optimization policy.
An example of this approach for an arbitrary customer is shown in \cref{appendix:opt_example}.
As with our CQL model this approach requires the offline policy evaluation technique described in \cref{section:ope} to estimate performance.

%% file: sections/results.tex
\section{Results}
\label{section:results}

\subsection{Online Auto Lending}

\subsubsection{Data Description}
\label{section:data}

For our first experiment, we use a static dataset on auto loan applications from an online lender in the United States provided by the Center for Pricing and Revenue Management.\footnote{\url{https://www8.gsb.columbia.edu/cprm/research/datasets}}
This dataset contains information on approximately 200,000 approved auto loan applications including the interest rate charged, the term of the loan, the approved amount, the FICO risk score, and whether the customer accepted the offer.
A description of the data, features, and pre-processing steps is available in \cref{appendix:data}. We also refer the reader to \citet{phillips2015effectiveness} for more detail on this dataset.

\subsubsection{The CQL Policy}

\begin{figure}[ht]%
    \vskip 0.15in
    \centering
    \subfloat[\centering Cumulative expected reward]{{\includegraphics[width=0.44\linewidth]{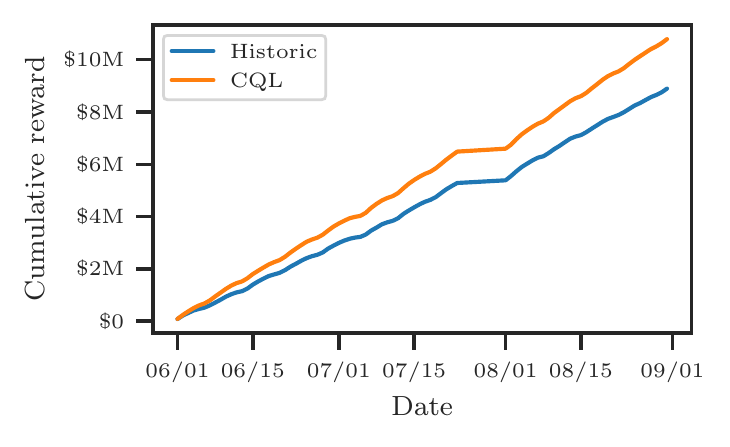}}\label{fig:cql_cumsum_return} }%
    \qquad
    \subfloat[\centering Price distribution]{{\includegraphics[width=0.44\linewidth]{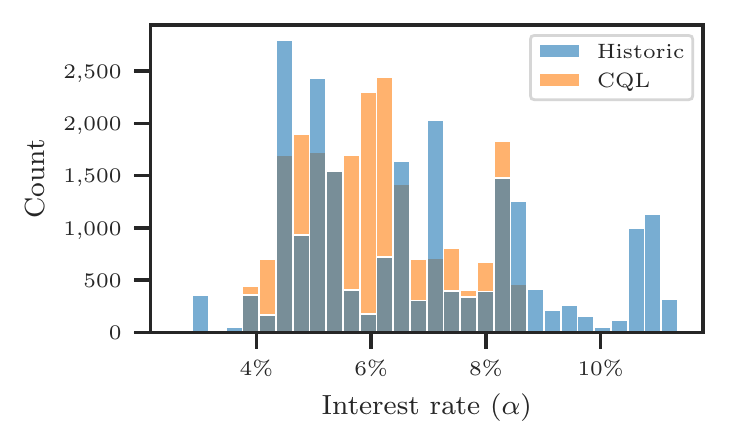}}\label{fig:cql_price_dist} }%
    \caption{Performance of the CQL policy on the test set for a single seed. Cumulative expected reward is estimated using the baseline logistic regression price-response model.}%
    \label{fig:cql_performance}%
    \vskip -0.1in
\end{figure}

We evaluate the performance of the CQL policy, $\picql$, on the test set using a logistic regression model to estimate the response probabilities, $p(Accept_{t}|s_{t}, a_{t})$, at the new prices and calculate expected reward.\footnote{See the Appendix for details on hyperparameters, train/val/test splits, and runtime as well as the estimated parameters of the logistic regression price-response model.}
Averaged over three seeds, our results indicate that the CQL agent is able to learn an effective policy that improves on historical pricing, $\pihist$, by approximately 21\% in expected profit while maintaining a less than 15\% mean absolute percentage deviation in prices from the existing policy.
\cref{fig:cql_price_dist} shows that despite the strong overlap between $\picql$ and $\pihist$, the new policy pushes the average price downwards from 6.8\% to 5.9\%.
This is consistent with the results of \citet{phillips2015effectiveness} who find evidence of historical over-pricing in this dataset due to a lack of subvented deals that were prevalent at many dealerships at the time which reduced the competitiveness of the lender's rates.

We also compare the CQL model against the profit-based optimization policy, $\piopt$, described in \cref{section:eval}.\footnote{
We restrict prices to be between 2.5\% to 12.5\% to speed up optimization, however, these constraints are not binding.}
This approach uses the same logistic regression price-response model used to evaluate performance but fitted on the training set to identify the price that maximizes the expected reward in \cref{eqn:reward} for each application in the test set.
With this policy we find an overall 34\% increase in expected profit which is associated with an average 24\% absolute percentage difference in price relative to the historical policy.
These results are consistent with the results of \citet{phillips2015effectiveness} and \citet{ban2021personalized} who estimate increases of 38\% and 47\% on the same auto loans dataset, respectively.
The differences in performance may be explained by the different test sets and the simplified reward functions they used that did not factor in credit risk or capital costs in the case of \citeauthor{ban2021personalized}.

\begin{figure}[ht]%
    \vskip 0.15in
    \centering
    \subfloat[\centering CQL]{{\includegraphics[width=0.44\linewidth]{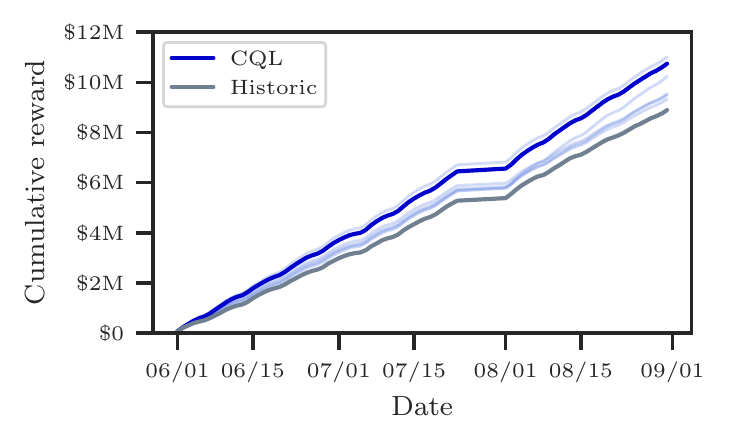} }}%
    \qquad
    \subfloat[\centering  Opt]{{\includegraphics[width=0.44\linewidth]{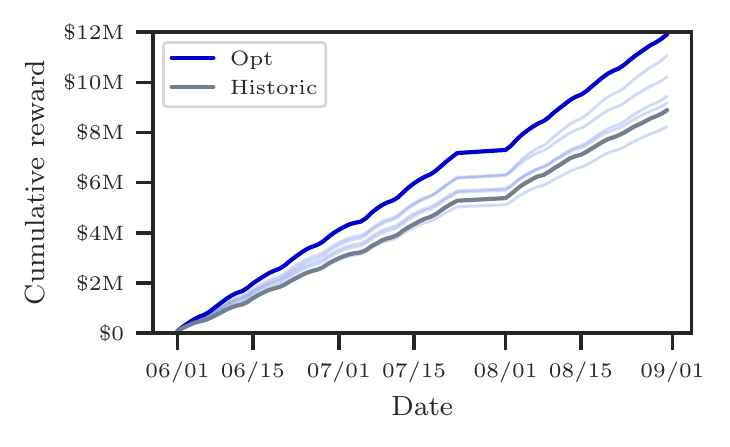} }}%
    \caption{Cumulative expected rewards over the test set for both the CQL and Opt policies evaluated using different price-response models with a single seed.
The dark blue line represents the estimated performance using the logistic regression price-response model and the lighter blue lines represent results using alternative price-response models.
These include parametric and non-parametric approaches such as gradient boosting, regularized logistic regression, and neural network models.}%
    \label{fig:elasticity_sensitivity}%
    \vskip -0.1in
\end{figure}

\cref{fig:elasticity_sensitivity} demonstrates the sensitivity of $\piopt$ performance to the assumed functional form of demand.
Using alternative non-parametric and parametric price-response models with comparable classification performance, we find high variance in $\piopt$ performance with estimates ranging from a -7\% decrease to a 34\% increase in total expected reward relative to $\pihist$ with an average estimate of 12.6\%.
There is also strong evidence of overfitting with the highest performance estimated by the logistic regression used to optimize the prices.
This sensitivity to the assumed functional form of demand is understated in previous literature and particularly problematic within this dataset due to the limited explainability of the feature set.
For example, the baseline logistic regression model only achieves a pseudo $R^{2}$ of 0.29 and AUC of 0.83.
In contrast, while $\picql$ still exhibits sensitivity to the price-response model used for evaluating performance, the estimates lie in a narrower range from 5\% to 24\% with an average of 13\%.
These results are consistent with the work of \citet{rana2014real} who test the effect of model misspecification on model-free reinforcement learning and parametric learning algorithms using synthetic data and find similar sensitivity of parametric methods that assume knowledge of demand.

\begin{figure}[ht]%
    \vskip 0.15in
    \centering
    \subfloat[\centering CQL]{{\includegraphics[width=0.44\linewidth]{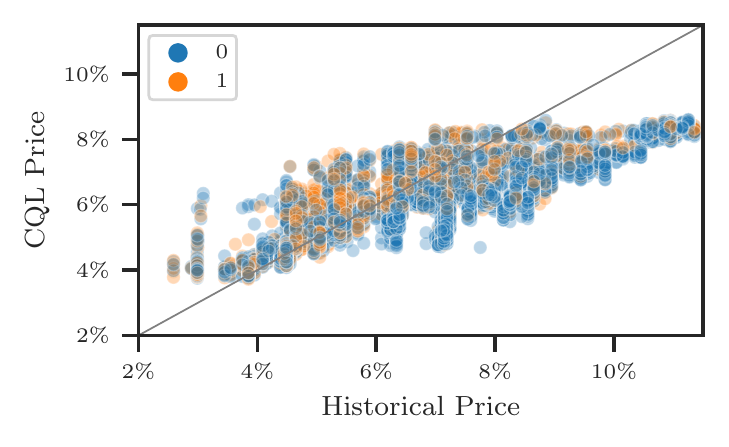} }}%
    \qquad
    \subfloat[\centering Opt]{{\includegraphics[width=0.44\linewidth]{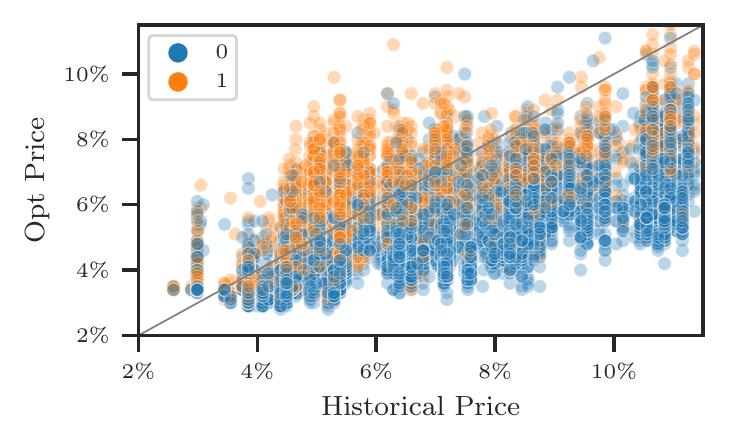} }}%
    \caption{Comparison of $\piopt$ and $\picql$ against $\pihist$ on the test set for a single seed. The blue points represent applications that did not accept the original price under $\pihist$ and the orange points represent loans that were funded.}%
    \label{fig:prices_scatter}%
    \vskip -0.1in
\end{figure}

The fact that $\picql$ achieves performance comparable to $\piopt$ is impressive given the relatively limited change in prices from the historical policy.
From \cref{fig:prices_scatter} we also observe that $\picql$ is able to implicitly estimate the sensitivity of applicants to prices.
Without knowledge of responses or rewards in the test set, it produces a policy that reduces prices for applicants who would have rejected the historical price by an average of 28\% while for applicants that would have accepted it, the algorithm maintains prices on average 95\% of what would have been quoted under $\pihist$.

In addition to the conservative improvements over the historic pricing policy, the benefits of the CQL algorithm also lie in its simplicity. 
Although we use estimates of the price-response curve and reward function for evaluating its policy, all that is required for training the CQL algorithm is a static dataset with features, offered prices, and rewards.
In contrast, a profit-based pricing approach often requires estimation of a separate price-response model, access to the reward function, and an optimization framework to price each application.

\subsubsection{The value of conservatism}

\begin{figure}[ht]%
    \vskip 0.15in
    \centering
    \subfloat[\centering Cumulative reward]{{\includegraphics[width=0.44\columnwidth]{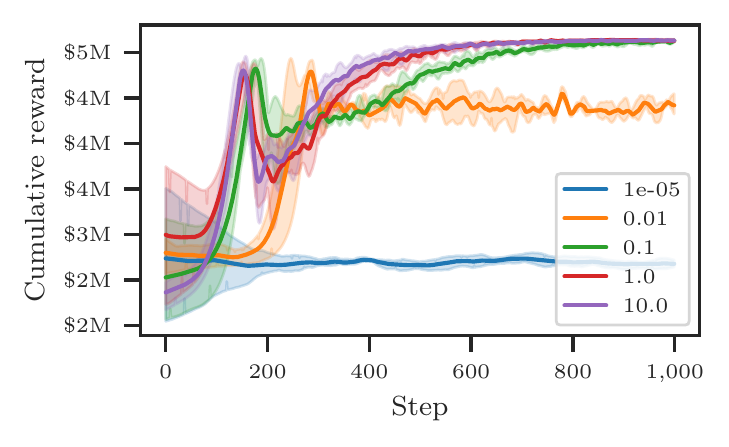} }}%
    \qquad
    \subfloat[\centering MAPD]{{\includegraphics[width=0.44\columnwidth]{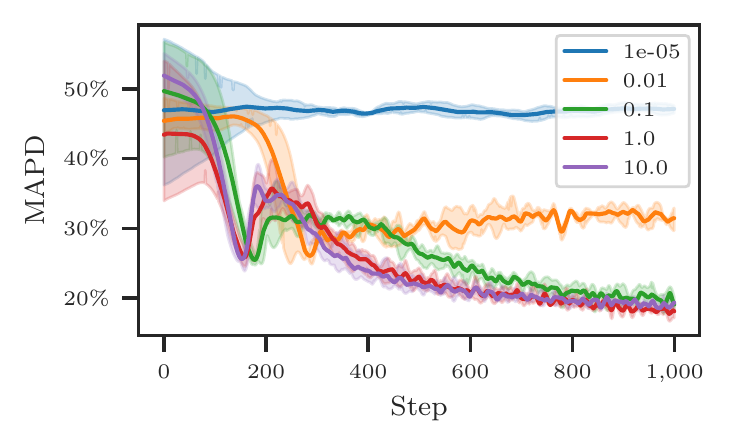} }     \label{fig:cql_conservative_mapd}%
}%
\caption{CQL performance on the first 10,000 test set applications as a function of training step with different fixed values of the trade-off parameter, $\alpha$, averaged over three seeds.
Cumulative reward is evaluated using the baseline price-response model described in \cref{section:ope}.}%
    \label{fig:cql_conservative}%
    \vskip -0.1in
\end{figure}

We assess the effect of distributional shift on performance by fixing different values of the trade-off parameter, $\alpha$, used to control the CQL regularization term in \cref{eqn:cql_loss}.
In the extreme case, where $\alpha$ is close to zero, we recover an offline version of the SAC algorithm.
In \cref{fig:cql_conservative} we observe that removing the penalty on out-of-distribution actions results in substantial variation in prices compared to the historical policy as a result of over-estimating the value of unseen state-action pairs.
This dramatically reduces performance and the algorithm is unable to learn an effective policy.
As we increase the value $\alpha$, we find that performance improves dramatically and the agent is able to learn stably.

\subsection{Synthetic Price Responses}
\label{section:synthetic-results}

For our second set of experiments, we create synthetic datasets from the historical auto loans data using various price-response models.
That is, we train a price-response model to estimate $p(\mli{Accept}_{t} | s_{t}, a_{t})$ on the historical dataset and use the predicted probabilities to sample from the Bernoulli distribution and replace the original accept decisions, $Accept_{t}$.
We isolate California which is the largest market and has approximately 30,000 applications but otherwise follow the same approach as in the previous sections.
We also compare our approach against an additional profit-based pricing model with a demand formula similar to the one introduced by \citet{ban2021personalized} that includes feature-dependent price effects (FDPE) by adding an interaction term for each of the features with price.

\subsubsection{Comparison against the optimal policy}
With access to the true price-response model, we are able to measure the performance of the different pricing policies against the optimal policy obtained by applying the profit-based optimization approach with the true model.
In \cref{table:synthetic-performance} we see the performance of the policies across different synthetic datasets.
When there is no demand misspecification, i.e., the profit-based model's price-response curve matches the dataset's, the profit-based pricing approaches are able to achieve near-perfect performance.
In real-world pricing applications, this is a strong assumption that is violated with the presence of non-stationarity, incomplete information, or complex demand behavior \citep{cheung2021hedging,luo2021distribution}.
For example, when we allow the coefficients of the price-response model to vary by customer segment or use a more expressive neural network model to represent demand behavior, the profit-based models are no longer able to achieve the same level of relative performance.
In these potentially more realistic scenarios, the CQL policy is able to achieve the same level of performance with approximately 50\% less price variation from the historical policy.
While this stability in prices does help mitigate the risks associated with demand misspecification, it limits the effectiveness of the CQL algorithm when large variations in price are required to achieve optimal performance.

\begin{table}[ht]
\vskip 0.15in
\centering
\caption{
Performance of the pricing policies on synthetic data created from five different price-response models averaged over three seeds.
$\pihist$, $\picql$, $\piopt$, and $\pioptfdpe$ refer to the historical policy, CQL policy, profit-based optimization policy, and profit-based optimization policy with feature dependent price effects, respectively. 
The \textit{logistic} dataset uses the same price-response model from $\piopt$ to generate the new accept decisions.
\textit{logistic-fdpe} includes interactions with price for each of the features. 
\textit{segmented} allows the coefficients of the price-response model to vary for customer segments assigned by an unsupervised Gaussian mixture model.
\textit{time-varying} allows the coefficients of the logistic regression in the test set to vary from the training set.
\textit{neural net} uses a deep neural network as the true price-response model.
MAPD is the mean absolute percentage deviation in price relative to the historical pricing policy, $\pihist$.
}
\label{table:synthetic-performance}
\resizebox{\linewidth}{!}{%
\begin{tabular}{l|ccc|cccc}
\toprule
\begin{tabular}[c]{@{}l@{}}\\\end{tabular} & \multicolumn{3}{c|}{\textbf{MAPD}} & \multicolumn{4}{c}{\textbf{\% of Optimal Return}} \\
\textbf{Dataset} &
\multicolumn{1}{c}{\textbf{$\picql$}} & 
\multicolumn{1}{c}{\textbf{$\piopt$}} &
\multicolumn{1}{c|}{\textbf{$\pioptfdpe$}} &
\multicolumn{1}{c}{\textbf{$\pihist$}} &
\multicolumn{1}{c}{\textbf{$\picql$}} &
\multicolumn{1}{c}{\textbf{$\piopt$}} &
\multicolumn{1}{c}{\textbf{$\pioptfdpe$}} \\
\midrule
\textbf{logistic} & 16.2\% & 24.3\% & 23.9\% & 74.5\% & 90.7\% & 99.8\% & 99.3\% \\
\textbf{logistic (FDPE)} & 16.0\% & 24.4\% & 24.8\% & 72.6\% & 87.2\% & 95.2\% & 98.9\% \\
\textbf{segmented} & 16.0\% & 25.5\% & 24.8\% & 57.4\% & 86.1\% & 83.4\% & 84.0\% \\
\textbf{time-varying} & 16.0\% & 25.6\% & 25.5\% & 88.1\% & 92.1\% & 91.2\% & 91.1\% \\
\textbf{neural net} & 16.1\% & 23.0\% & 26.3\% & 76.9\% & 84.0\% & 83.1\% & 82.9\% \\
\bottomrule
\end{tabular}
}
\vskip -0.1in
\end{table}

\subsection{Ethical considerations}

Several key issues should be considered before deploying a pricing policy based on our proposed method.
In particular, our pricing agent is trained on a dataset of historical pricing decisions and may learn to recreate any biases present within that data.
This may need to be addressed with additional constraints or appropriate data cleansing.
Care must also be taken when selecting features such that they do not unfairly target protected groups either directly or by proxying for protected characteristics.

%% file: sections/conclusion.tex
\section{Conclusion and Future Work}
\label{section:conclusion}

In this paper, we propose a model-free offline reinforcement learning approach to pricing consumer credit. 
This approach makes no assumptions on the functional form of demand and introduces a formulation in which actions may impact the future state of the environment and rewards.
We also demonstrate using synthetic and real data that this approach is able to improve on the existing pricing policy while showing robustness to misspecification of the underlying demand behaviour.

One logical extension of our work is to consumer goods as well as other types of consumer debt products.
Future work may also seek to extend the action space to include the underwriting decision as well as incorporate capital and maximum portfolio risk requirements.
In addition, as the rewards are not realized until the end of the loan term
another challenge may be to develop an approach allowing for future customer behavior to affect a previously learned policy.  

%% file: sections/acknowledgements.tex
\section*{Acknowledgements}

Raad Khraishi's research is currently funded by NatWest Group.
We thank Greig Cowan, Graham Smith, and Zachery Anderson for their valuable feedback and support.
We would also like to thank Devesh Batra for his feedback on earlier drafts.

%% file: sections/appendix.tex
\section{Auto Lending Dataset}
\label{appendix:data}

We use the period from July 2002 to August 2004 which is the latest month in which all offers expire before the end of the dataset.
In addition, we exclude \textless 1,000 records with anomalous loan values less than \$5,000, missing values for state, or duplicate rows across ApproveDate, ApplyDate, Tier, FICO, CarType, State, LoanType, Rate, Amount, and Term.

\subsection{Data dictionary}
\label{appendix:data-dictionary}
\begin{center}
\begin{table}[h!]
\centering
\caption{\label{tab:data-dictionary}Auto lending data dictionary}
\begin{tabular}{@{}ll@{}}
\toprule
\textbf{Field} & \textbf{Definition}                                          \\ \midrule
Accept              & 1 for Funded, 0 for Non-funded                               \\
Rate                & Customer rate                                                \\
FICO                & Fico score                                                   \\
Tier                & Segmentation based on fico scores                            \\
State               & Customer state                                               \\
Type                & Finance/Refinance                                            \\
ApplyDate           & Application date                                             \\
ApproveDate         & Approval date                                                \\
DaysSinceApp        & Days between application date and approval date              \\
Term                & Approved term                                                \\
Amount              & Loan amount approved                                         \\
PreviousRate        & Previous interest rate for a refinanced car                  \\
CarType             & New, Used or Refinanced                                      \\
CompetitionRate     & Competitor's rate                                            \\
PrimeRate           & Prime rate                                                   \\
Months              & Month indicator                                              \\
TermClass           & Segmentation based on terms                                  \\
PartnerBin          & Segmentation based on partners                               \\
DayOfWeek           & Day of week                                                  \\
MonthOfYear         & Month of year                                                \\ \bottomrule
\end{tabular}
\end{table}
\end{center}

\subsection{Train/val/test split}
\label{appendix:train-split}
\begin{center}
\begin{table}[ht]
\centering
\caption{\label{tab:train-split}Train/val/test split. To assess out-of-sample performance we split our dataset into training, validation, and testing sets.
For hyperparameter selection we use data from May 2004 which is added to the training set before assessing final performance on the test set.}
\begin{tabular}{@{}lllrrr@{}}
\toprule
\textbf{Dataset} & \textbf{Start} & \textbf{End} & \multicolumn{1}{l}{\textbf{Observations}} & \multicolumn{1}{l}{\textbf{\% Accept}} & \multicolumn{1}{l}{\textbf{Avg. Reward}} \\ \midrule
Train   & 2002-07-01  & 2004-04-30   & 161,314  & 21.0\%  & \$324 \\
Val     & 2004-05-01  & 2004-05-31   & 6,482    & 26.1\%  & \$447 \\
Test    & 2004-06-01  & 2004-08-31   & 21,471   & 25.7\%  & \$417 \\ \bottomrule
\end{tabular}
\end{table}
\end{center}

\newpage

\section{Reproducability}
\label{appendix:reproducability}

\subsection{CQL Hyperparameters}
\label{appendix:hyperparams}
\begin{center}
\begin{table}[ht!]
\centering
\caption{\label{tab:hyperparams}CQL Hyperparameters.
We mainly use the default parameters from \lib{d3rlpy} with the exception of the hidden layers, discount rate, dropout, weight decay, and min-max scaling of the actions, features, and rewards.
The hidden layers and discount rate were tuned on the validation set described in \cref{appendix:train-split}.
For the synthetic results in \cref{section:synthetic-results}, we use the same hyperparameters but with only two hidden layers.
}
\begin{tabular}{ll}
\toprule
n\_epochs             &       20 \\
batch\_size           &      256 \\
hidden\_units         &       [64, 64, 64, 64] \\
n\_steps              &        1 \\
weight\_decay         &   0.0001 \\
gamma                &    0.999 \\
alpha\_threshold      &       10 \\
conservative\_weight  &        5 \\
dropout\_rate         &      0.2 \\
n\_critics            &        2 \\
use\_batch\_norm       &    False \\
action\_scaler        &  'min\_max' \\
scaler               &  'min\_max' \\
reward\_scaler        &  'min\_max' \\
n\_action\_samples     &       10 \\
actor\_learning\_rate  &   0.0001 \\
critic\_learning\_rate &   0.0003 \\
temp\_learning\_rate   &   0.0001 \\
alpha\_learning\_rate  &   0.0001 \\
initial\_alpha        &        1 \\
q\_func\_factory       &     'mean' \\
\bottomrule
\end{tabular}
\end{table}
\end{center}

\subsection{Runtime}
Training the algorithm on the full dataset for 20 epochs takes approximately one hour running on CPU on a 2020 MacBook Pro with a 2 GHz Quad-Core Intel Core i5 processor while a single action prediction takes approximately 640 microseconds.

\subsection{Seeds}
We use the following seeds in order when running our experiments: 333, 42, and 3.

\newpage

\section{Baseline price-response model}
\label{section:baseline-price-response}
\begin{center}
\begin{tabular}{lclc}
\toprule
\textbf{Dep. Variable:}             &      Accept       & \textbf{  No. Observations:  } &   189267    \\
\textbf{Model:}                     &      Logit       & \textbf{  Df Residuals:      } &   189252    \\
\textbf{Method:}                    &       MLE        & \textbf{  Df Model:          } &       14    \\
\textbf{Date:}                      & Sun, 09 Jan 2022 & \textbf{  Pseudo R-squ.:     } &   0.2946    \\
\textbf{Time:}                      &     14:09:56     & \textbf{  Log-Likelihood:    } &   -69912.   \\
\textbf{converged:}                 &       True       & \textbf{  LL-Null:           } &   -99105.   \\
\textbf{Covariance Type:}           &    nonrobust     & \textbf{  LLR p-value:       } &    0.000    \\
\bottomrule
\end{tabular}
\begin{tabular}{lcccccc}
                                    & \textbf{coef} & \textbf{std err} & \textbf{z} & \textbf{P$> |$z$|$} & \textbf{[0.025} & \textbf{0.975]}  \\
\midrule
\textbf{Intercept}                  &      44.0661  &        1.632     &    26.996  &         0.000        &       40.867    &       47.265     \\
\textbf{C(CarType)[T.R]}    &       0.2901  &        0.046     &     6.242  &         0.000        &        0.199    &        0.381     \\
\textbf{C(CarType)[T.U]}    &       2.4328  &        0.021     &   116.143  &         0.000        &        2.392    &        2.474     \\
\textbf{C(PartnerBin)[T.2]} &      -1.1915  &        0.024     &   -49.851  &         0.000        &       -1.238    &       -1.145     \\
\textbf{C(PartnerBin)[T.3]} &      -0.3275  &        0.014     &   -22.728  &         0.000        &       -0.356    &       -0.299     \\
\textbf{C(Tier)[T.2]}       &      -0.1404  &        0.025     &    -5.710  &         0.000        &       -0.189    &       -0.092     \\
\textbf{C(Tier)[T.3]}       &      -0.0351  &        0.034     &    -1.036  &         0.300        &       -0.101    &        0.031     \\
\textbf{C(Tier)[T.7]}       &       0.3245  &        0.053     &     6.145  &         0.000        &        0.221    &        0.428     \\
\textbf{rate}                       &      -0.6599  &        0.010     &   -62.882  &         0.000        &       -0.680    &       -0.639     \\
\textbf{PrimeRate}                   &       0.8124  &        0.047     &    17.402  &         0.000        &        0.721    &        0.904     \\
\textbf{CompetitionRate}          &       0.1619  &        0.024     &     6.744  &         0.000        &        0.115    &        0.209     \\
\textbf{PreviousRate}     &       0.0021  &     4.78e-05     &    44.980  &         0.000        &        0.002    &        0.002     \\
\textbf{np.log(Amount)}   &      -1.7937  &        0.019     &   -94.566  &         0.000        &       -1.831    &       -1.756     \\
\textbf{np.log(FICO)}      &      -4.4786  &        0.243     &   -18.429  &         0.000        &       -4.955    &       -4.002     \\
\textbf{Term}                       &       0.0518  &        0.001     &    52.238  &         0.000        &        0.050    &        0.054     \\
\bottomrule
\end{tabular}
\end{center}

\newpage

\section{Profit-based pricing example}
\label{appendix:opt_example}
\begin{figure}[ht]%
    \vskip 0.15in
    \centering
    \subfloat[\centering Price-response curve]{{\includegraphics[width=0.44\columnwidth]{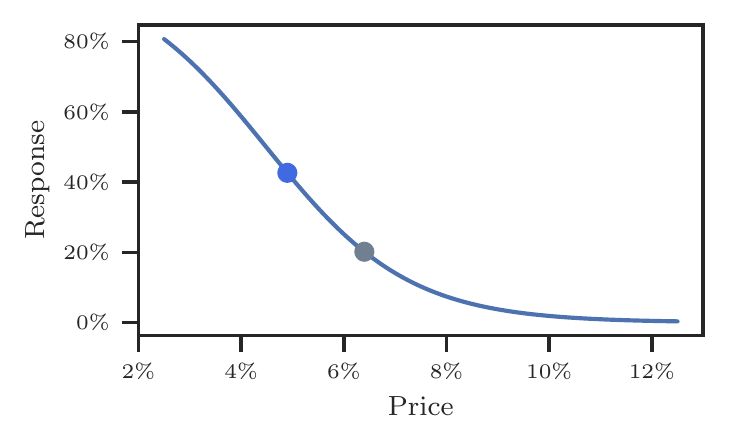} }}%
    \qquad
    \subfloat[\centering Expected reward]{{\includegraphics[width=0.44\columnwidth]{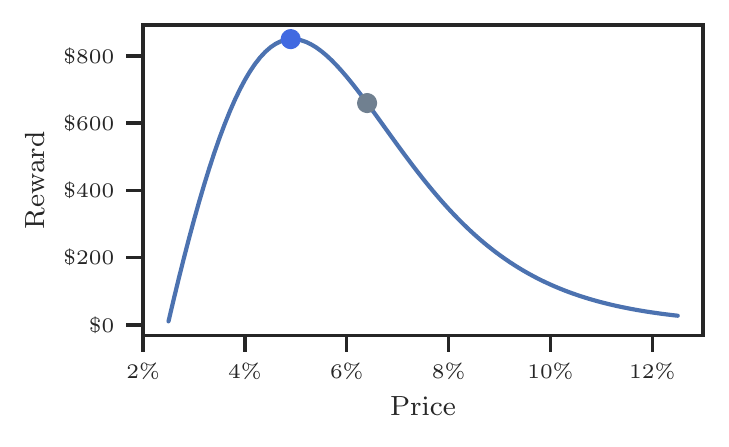} }}%
    \caption{Profit-based price optimization for a single arbitrary customer. The plot on the left shows the probability of a customer accepting a loan at different prices estimated using logistic regression. The plot on the right uses these probabilities to estimate the expected reward at each price point. The grey dot represents the historical price and the blue dot represents the reward maximizing price.}%
    \label{fig:opt_example}%
    \vskip -0.1in
\end{figure}
\newpage